\documentclass[10pt,twocolumn,letterpaper]{article}

\usepackage{wacv}
\usepackage{times}
\usepackage{epsfig}
\usepackage{graphicx}
\usepackage{amsmath}
\usepackage{amssymb}

\usepackage{algorithm}
\usepackage{algcompatible}
\usepackage{algpseudocode}
\graphicspath{{img/}}

\usepackage{xcolor}
\usepackage{dblfloatfix}

\newcommand\numberthis{\addtocounter{equation}{1}\tag{\theequation}}
\newcommand{\qdr}{\textit{Quick-draw}}
\newcommand{\xent}{\textsc{xent}}
\newcommand{\mwx}{\textsc{mwx}}

\usepackage[pagebackref=true,breaklinks=true,letterpaper=true,colorlinks,bookmarks=false]{hyperref}

\wacvfinalcopy 

\def\wacvPaperID{400} 

\ifwacvfinal\fi
\begin{document}

\title{Can I teach a robot to replicate a line art}

\author{B.V. Raghav \\
Indian Institute of Technology Kanpur\\
{\tt\small bvraghav@iitk.ac.in}
\and
Subham Kumar \\
Indian Institute of Technology Kanpur\\
{\tt\small subhamkr@iitk.ac.in}
\and
Vinay P. Namboodiri \\
Indian Institute of Technology Kanpur\\
{\tt\small vinaypn@iitk.ac.in}
}

\maketitle
\ifwacvfinal\thispagestyle{empty}\fi

\begin{abstract}

  Line art is arguably one of the fundamental and versatile modes of
  expression. We propose a pipeline for a robot to look at a grayscale
  line art and redraw it. The key novel elements of our pipeline are:
  a) we propose a novel task of mimicking line drawings, b) to solve
  the pipeline we modify the Quick-draw dataset to obtain supervised
  train- ing for converting a line drawing into a series of strokes c)
  we propose a multi-stage segmentation and graph interpretation
  pipeline for solving the problem. The resultant method has also been
  deployed on a CNC plotter as well as a robotic arm. We have trained
  several variations of the proposed methods and evaluate these on a
  dataset obtained from Quick-draw. Through the best methods we
  observe an accuracy of around 98\% for this task, which is a
  significant improvement over the baseline architecture we adapted
  from. This therefore allows for deployment of the method on robots
  for replicating line art in a reliable manner. We also show that
  while the rule-based vectorization methods do suffice for simple
  drawings, it fails for more complicated sketches, unlike our method
  which generalizes well to more complicated distributions.

\end{abstract}

\section{INTRODUCTION}
\label{sec:introduction}

\begin{figure*}[b]
  \centering
  \includegraphics[width=1.0\linewidth]{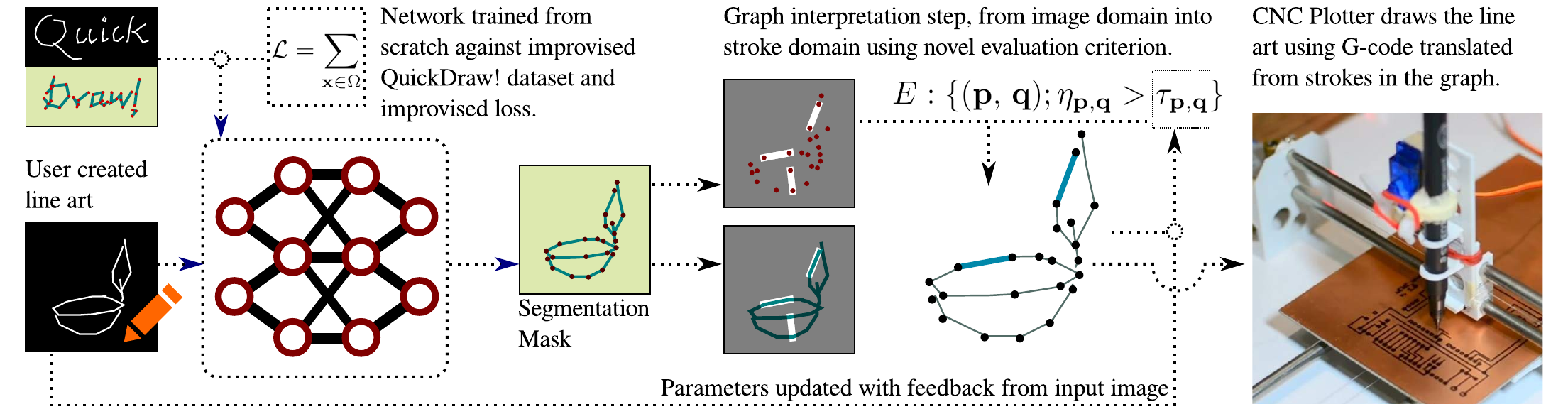}
  \caption{Left to Right. An artist drawing; digital copy of the
    drawing; segmented version; drawing by a \textsc{cnc} plotter.}
  \label{fig:teaser}
\end{figure*}

Line is a fundamental building block in art form, and so has been
emphasised by pioneers of historical art movements led by the likes of
Picasso, Paul Klee, Kandinsky and Piet Mondrian in the past. In modern
times drawing as an activity has been studied in the context of
problem solving in a design process, while sampling mostly over line
art\cite{ullman_importance_1990,visser2006cognitive}.

There is one question that forms the under-current of our
investigation here,
\begin{quotation}
  If I provide \textit{any} line drawing to a robot, how can I teach
  it to \textit{look at it,} and draw the same on a paper using a pen.
\end{quotation}
This is a different challenge that has to the best of our knowledge
not been addressed previously. There are vectorization based methods
in Computer Graphics that aim to convert an input sketch into
strokes. However, these would lose salient stroke information and
would not preserve the line drawing (refer
\S~\ref{sec:generalization}). Other approaches aim to parse sketch
into semantic segments. Again, this would not be useful for obtaining
stroke information. In contrast to these, we aim to {\it replicate}
the line drawing by converting an input raster line drawing image into
a set of strokes (refer \S~\ref{sec:methodology}) and the
corresponding instructions (refer \S~\ref{sec:graph-partition}) that
can be used by an automated system. To obtain this, we present an
approach different from the popular vectorization
models~\cite{favreau2016fidelity,tombre_vectorization_2000}, the
segmentation
models~\cite{Schneider:2016:ESS:2965650.2898351,kaiyrbekov_stroke-based_2019},
the interesting developments of the drawing
robots~\cite{galea_tethered_2017,tresset_portrait_2013} and further
present its applicability.

We treat this problem as a special case of vectorization, by learning
a function to map a line drawing from image domain to a graph
structure. To this effect, we deploy, as a prior step in the pipeline,
an image to image translation to facilitate the inference, using a
deep segmentation neural network. There is also a post process step
involved, to retrieve a sequence of strokes from the graph
structure. This is further translated to \textsc{gcode} to input to a
\textsc{cnc} plotter (refer \S~\ref{sec:graph-partition}), or used
directly by factory software to compute trajectory for a robotic arm.

We propose a novel \textit{U-Net based deep neural network}
\textsc{(dnn)} to implement the image to image translation, trained
over a novel dataset built upon the \qdr{}
dataset~\cite{ha_neural_2017}, and through a tailor-made training
objective. The novelty in the dataset lies in generating on the fly,
the segmentation true labels.

We propose a simple yet effective method, to infer the strokes as a
graph data-structure, from the image domain. We utilise the
segmentation of the image into \textit{corners} channel, which for
most practical cases have one to one correspondence with vertices of
the graph, and \textit{lines}, that show a similar behaviour of
edges. The inference is made using an iterative update to a
parametrized filter criterion, that selects a subset of the edges
as graph proposal after every update.

To extract stroke sequences from the graph structure, we effectively
utilise a recursive approach and illustrate the utility using two
devices, namely \textsc{cnc} plotter, and a robotic arm.

To conclude through this paper we make the following main contributions:
\begin{itemize}
\item Propose a multistage pipeline, for segmentation, and graph
  interpretation;
\item Novel architecture for the deep segmentation network;
\item Annotated dataset for training the network;
\item Improvised loss function as the training objective; and
\item Feedback based iterative update to infer graph structure from
  image domain.
\end{itemize}

\section{RELEVANT WORKS}
\label{sec:relevant}

\begin{figure*}[b]
  \centering
  \includegraphics[width=0.9\textwidth]{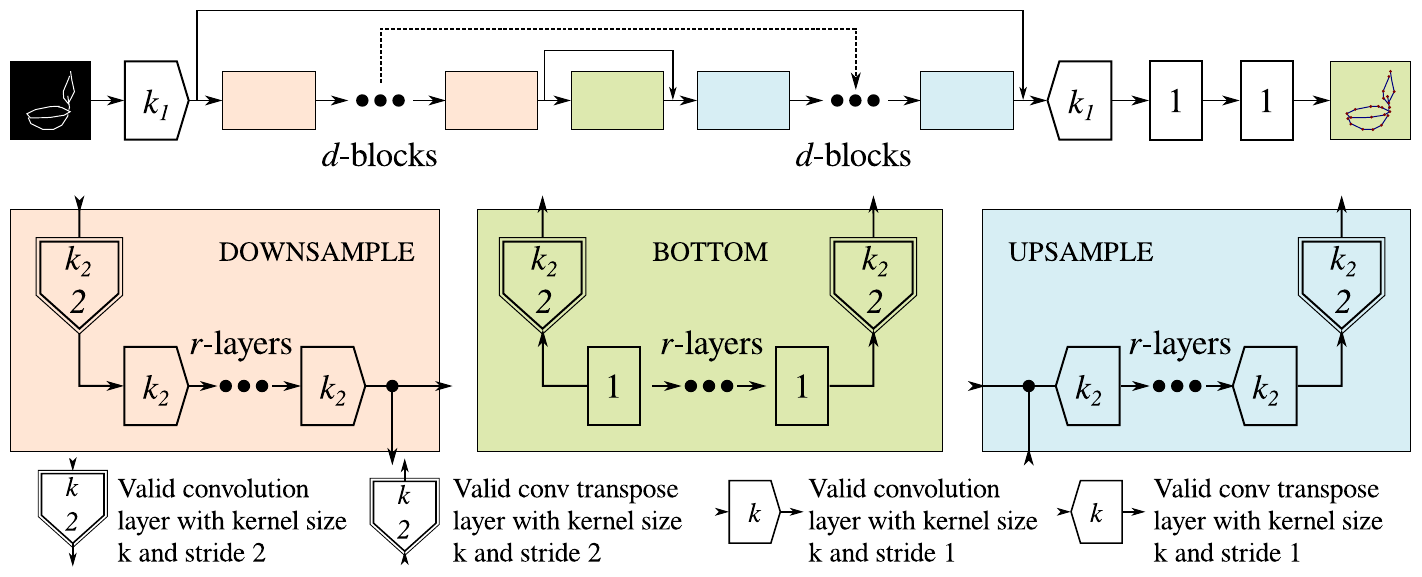}
  \caption{Reinterpretation of the U-Net architecture with modular
    blocks for \textit{downsampling}, \textit{upsampling} and
    \textit{bottom}.}
  \label{network}
\end{figure*}

Research in robotics has seen some recent developments, while trying
to create a drawing robot. Paul, the robotic arm, was trained to
sketch a portrait, with inherent inputs from an artist, while
deploying a stack of classical vision methods into their
pipeline~\cite{tresset_portrait_2013}. More recently,
~\cite{galea_tethered_2017, galea_stippling_2016} investigated whether
a similar feat can be achieved through a quad rotor, with the aim of
``applying ink to paper with aerial robots.''  Their contributions
include computing a stipple pattern for an image, a greedy path
planning, a model for strategically replacing ink, and a technique for
dynamically adjusting future stipples based on past errors.
Primarily, such efforts are based on space filling exercise, and focus
on the control systems, rather than the accuracy of the input
image. Our investigation deals more with capturing the line nature of
artwork, and is a different exercise altogether.

Robots with high quality manipulators~\cite{song_artistic_2018} have
been used to draw line based art on uneven surface(s), exploiting the
availability of impedance control. Earlier, Fu et
al.~\cite{fu_animated_2011} illustrated an effective use of graph
minimisation over hand crafted features to predict a reasonable
ordering of given strokes.  However, these methods rely on vector
input. We on the other hand propose a pipeline that reads from a
grayscale image.

Sketch vectorization is very close to the graph structured learning
part of our pipeline, but it has its own set of inherent
problems~\cite{tombre_vectorization_2000}. Two specific approaches
include skeleton based methods, that have to deal with corner
artefacts, and contour based methods, that deal with problems related
to one-to-one correspondences. Recent developments deal with noise
removal and Favreau et.al.~\cite{favreau2016fidelity} in particular,
investigate a higher order problem of fidelity vs simplicity of the
method. Inherent to these approaches is the loss of information from
the image domain in order to mitigate the basic problems. We propose
to use segmentation techniques which are comparatively better at
preserve the information as shown further in
\S~\ref{sec:generalization}.

Segmentation and labelling of sketches has been subject to profound
research in graphics and vision. Schneider
\etal~\cite{Schneider:Tuytelaars:2014} had successfully used an
ensemble of Gaussian Mixture Models and Support Vector Machines with
Fisher Vectors as distance metric to discriminate strokes; Further, it
was illustrated that use of Conditional Random Fields over a
relationship graph described by proximity and enclosure between
the strokes, performs well in assigning semantic labels to
strokes~\cite{Schneider:2016:ESS:2965650.2898351}. The classical
methods use hard coded features like SIFT, that was used here, as the
basic building block for learning algorithms.

Recently, convolutional neural networks have been proven to be highly
effective in classifying segmented strokes as given
labels~\cite{cheng_part-level_2018}.  Li \etal~\cite{li_fast_2018}
have illustrated the effectiveness of U-Net based deep neural network
architecture in the context of semantic segmentation, with the help of
3D objects dataset. The problems involving semantic segmentation
whether used in classical context or with the deep learning based
methods, target a higher level problem, namely semantics. We formulate
our problem at a much lower level, by distinguishing corners from
lines. Some problems also exploit the characteristics of dataset, for
example data synthesised out of 3D objects, used by Li
\etal~\cite{li_fast_2018}, has a much stronger sense of enclosure,
which are not necessarily true for the hand drawn counterparts, that
form our dataset.

Sequence to sequence translation using deep learning methods have
shown to be successfully deployed both as a discriminative model
\cite{wu_sketchsegnet_2018} as well as a generative model in the
context of sketch
segmentation~\cite{kaiyrbekov_stroke-based_2019}. The generative model
has been used to reconstruct a sketched symbol. Sequences do make
sense to work in stroke domain, where a sequence of vector input is
available to begin with. Our work deals with an image as in input.

 
\section{PIPELINE}
\label{sec:pipeline}

The pipeline as shown in Fig.~\ref{fig:teaser} can be split to three
significant parts, namely

\paragraph{Image to Image Translation}
The first part takes the raw sketch, a grayscale image as an input,
and translates it into a segmentation mask, one for each label, which
in our case are \textit{background, corners,} and \textit{lines}.

We implement this using a \textsc{dnn}, a variant of U-Net, that
avoids cropping the images, and retains the valid convolution
operations. This has been further detailed out in
\S~\ref{sec:inference} and \S~\ref{sec:loss_function}.

\paragraph{Graph Interpretation}
The second part uses the segmentation channels of lines and corners as
input and infers a graph structure, so that its vertices of the graph
represent the cusp or end points, and its edges represent the curve
drawn between them.

This is implemented using three parts; first is the inference of
vertices, using connected component analysis of corners channel;
second is a plausibility score of a given pair of vertices, evaluated
using the lines channel; and finally a filter criterion based on a per
edge threshold, that gets iteratively updated using feedback from the
input image. This has further been detailed out in
\S~\ref{sec:graph-interp}.

\paragraph{Utility}
The last part takes the graph structure and partitions into a
sequence of strokes using a recursive approach. We further detail out
the algorithm in \S~\ref{sec:graph-partition}.


\section{METHODOLOGY}
\label{sec:methodology}

\subsection{Discovering the segmentation network}
\label{sec:inference}

U-Net, proposed by Ronnenerger~\cite{ronneberger_u-net_2015}, was an
extension of fully convolutional
network~\cite{shelhamer_fully_2017,long_fully_2015}, and was shown to
be highly effective in the context of biomedical segmentation.
Compared to the FCN, U-Net proposed a symmetric architecture, to
propagate the skip connections, within the convolutional network. The
effectiveness in the context of semantic segmentation was also shown
by Li \etal~\cite{li_fast_2018}

Additionally, since downsample/upsample is followed by valid
convolutions, U-Net forwards only a crop of the activations as a skip
connection. Consequently, the output of the vanilla U-Net produces an
output image with size smaller than that of the input image,
\(H_{\mathrm{out}}, W_{\mathrm{out}} < H_{\mathrm{in}},
W_{\mathrm{in}}\). As a counter measure, the training and test images
were augmented by tiled repetition at borders, for the vanilla U-Net.

We on the other hand create an exact mirror of operations in the U-Net
architecture, i.e. downsampling is followed by valid convolutions and
upsampling follows valid transpose convolutions. This eliminates the
need for a tiling-based augmentation, and further gives way to nice
convolution arithmetic.

Yu et.al \cite{yu_sketch--net_2017} observed that in the context of
sketches, the network is highly sensitive to the size of ``first layer
filters'', the ``larger'' the better. Although they used a kernel size
of \((15\times{}15)\) for the first layer, we found that a size as
low as $7\times{}7$ is effective in our case.

Another notable difference, inspired by Radford et. al's observations
~\cite{radford_unsupervised_2015}, has been in the downsampling and
upsampling operations. U-Net uses ``\emph{$2\times2$ max pooling operation
  with stride 2 for downsampling}'' [\textsc{and} \emph{a
  corresponding upsampling strategy}.]  We have, on the other hand,
used valid strided convolutions for downlasmpling, and valid strided
transpose convolutions for up-sampling.

Our network, as shown in the Fig.~\ref{network} is
parameterised with a 4-tuple,\((k_1\, k_2\, d\, r)\), where
\begin{itemize}
\item \(k_1\) is the kernel $(k_1 \times k_1)$ of the first convolution
  layer.
\item \(k_2\) is the common kernel $(k_2 \times k_2)$ of all
  subsequent layers.
\item \(d\) is the depth of the U-Net, that is to say number of times
  to downsample during the contracting phase.
\item \(r\) is the number of horizontal layers along each depth level
\end{itemize}

\subsection{Training Objective}
\label{sec:loss_function}

\begin{figure}[!bp]
  \centering
  \includegraphics[width=\columnwidth]{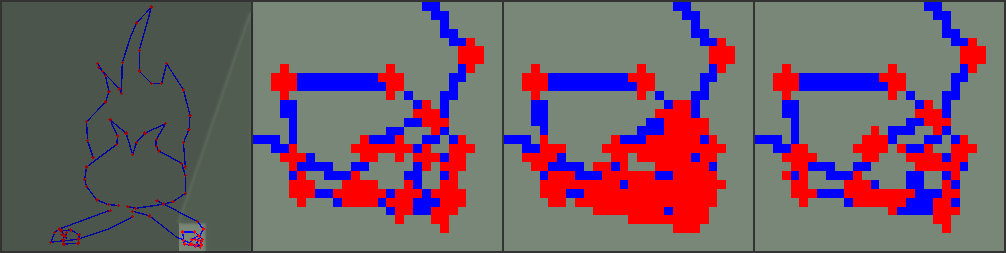}
  \caption{Effect of the change in weights for cross entropy
    loss. \textit{Left to Right.} true labels highlighting selected
    area; blown up version of selected area in true labels; prediction
    in selected area using $\omega(\ell(\mathbf{x}))$; using max
    weights.}
  \label{fig:xent-effect}
\end{figure}

We started with the standard, weighted softmax cross entropy loss, as
in the case of U-Net. For each pixel in the image,
$\mathbf{x} \in \Omega; \Omega \subset \mathbb{Z}^2$, if the network
activation corresponding to a given label $i$, be expressed as,
$a_i(\mathbf{x});\,\, 0 < i \leqslant K$, we express the predicted
probability of the label $i$, as the softmax of the activation,
$p_i(\mathbf{x}) = \mathrm{exp}(a_i(\mathbf{x})) /
\sum_j^K\mathrm{exp}(a_j(\mathbf{x}))$. We want
$p_{\ell(\mathbf{x})} \approx 1$ and
$p_{i}(\mathbf{x}) \approx 0; i \neq \ell(\mathbf{x})$ for the true
label $\ell(\mathbf{x})$. So we penalise it against the weighted cross
entropy, as follows:

\begin{align}
  \label{eq:xent}
  \mathcal{L} &= \sum_{\mathbf{x} \in \Omega} w(\mathbf{x})
                \log(p_{\ell(\mathbf{x})}(\mathbf{x}))
\end{align}

where $w(\mathbf{x})$ represents the weight map; which was initialised
inversely proportional to class frequencies,
$w(\mathbf{x})~\gets~\omega(\ell(\mathbf{x}));~\omega:\Lambda\to\mathbb{R}$,
where $\Lambda$ is the set of labels.

Further, we investigated a variant using max weights,
$w(\mathbf{x})\gets\omega(\max(\ell(\mathbf{x}), \rho(\mathbf{x})))$,
where $\rho(\mathbf{x})$ is the predicted label. This significantly
improves the result, as shown in Fig.~\ref{fig:xent-effect}. Even
the prediction of very complicated structures resembles the true
distribution.


\subsection{Graph Interpretation}
\label{sec:graph-interp}

There are two key observations in interpreting the graph $G(V,E)$
representing the strokes, from a segmentation mask. Firstly,
\textit{each corner is a set of strongly connected components in the
  channel}, so we collect their centroids into the set of vertices,
$V: \{\mathrm{centroid}(i); 0<i\leqslant{}N\}; V \subset
\mathbb{R}^{2}$, where N is the number of connected components in the
corners channel.


Secondly, \textit{along the line connecting centroids there is a peak
  response from lines channel,} $Y_{\mathrm{lines}}$. We define
\textit{a plausibility score}, $\eta_{\mathbf{p}, \mathbf{q}}$ of an
edge between an arbitrary pair of vertices,
$\mathbf{p}, \mathbf{q} \in V; \mathbf{p} \neq \mathbf{q}$, as the
average value of the lines channel masked by a region of interest
(\textsc{roi}) $M$ between them.  $M$ is defined as a narrow
rectangular region, of width $\beta$ centred along the line segment
$\overline{\mathbf{pq}}$. We define a per-pair, threshold parameter
$\tau_{\mathbf{p},\mathbf{q}}$ based filter criterion,
$C_{\mathbf{p},\mathbf{q}}$ as,

\begin{align*}
  C_{\mathbf{p},\mathbf{q}} &= \mathbf{I}\left[\eta_{\mathbf{p},\mathbf{q}} >
                              \tau_{\mathbf{p},\mathbf{q}}\right]\numberthis
                              \label{eq:criterion} \\
  \eta_{\mathbf{p},\mathbf{q}} &= \frac{1}{|M|}\sum_{\mathbf{x} \in
                                 M}Y_{\mathrm{lines}}(\mathbf{x})
\end{align*}
where, $\mathbf{I}$ is the indicator function.

Our experiments reveal that the threshold parameter varies with vertex
pairs. We adapt $\tau_{\mathbf{p},\mathbf{q}}$, using a feedback loop,
triggered by analysing the difference between the rendered strokes and
the input image, using connected component analysis. The blobs thus
obtained are characterised as \textit{a) absent blobs}, that are
present in input, but not rendered,
$\mathbb{B^+}:\{\textrm{bounding box} \mathrel{}\forall\mathrel{}
\textrm{absent blobs}\}$; and \textit{b) superfluous blobs,} that are
rendered, but not present in input,
$\mathbb{B^-}:\{\textrm{bounding box} \mathrel{}\forall\mathrel{}
\textrm{superfluous blobs}\}$. We do an iterative update of threshold
parameter for all pairs of vertices as:

\begin{align*}
  \tau^{i+1}_{\mathbf{p},\mathbf{q}} &=
                                       \tau^{i}_{\mathbf{p},\mathbf{q}} (1 + \lambda\,\delta_{\mathbf{p},\mathbf{q}}) \numberthis\label{eq:update} \\
  \delta_{\mathbf{p},\mathbf{q}} &= \begin{cases} 
    -1, &
    \text{if}\mathrel{}\exists\mathrel{}i\mathrel{}\textrm{st.}\mathrel{}
    (\mathbf{p},\mathbf{q})\mathrel{}\textrm{inside}\mathrel{}B^+_i\in \mathbb{B^+}; \\
    1, &
    \text{if}\mathrel{}\exists\mathrel{}j\mathrel{}\textrm{st.}\mathrel{}
    (\mathbf{p},\mathbf{q})\mathrel{}\text{inside}\mathrel{}B^-_j\in \mathbb{B^-}; \\
    0, & \textrm{otherwise}
   \end{cases}
\end{align*}
where $\lambda$ is an update hyperparameter.


\subsection{Strokes and Gcode}
\label{sec:graph-partition}

In order to compute the strokes, we follow a recursive approach, so that
every edge that is traversed is popped out of the graph, and
revisiting the vertex is allowed. See
Algorithms~\ref{algo:get_sequence},~\ref{algo:strokes_gen}. 

\begin{algorithm}
  \caption{\texttt{Get\_Sequence} Recursive Procedure for generating
    Sequence of a Stroke}
  \label{algo:get_sequence}
  \begin{algorithmic}[1]
  
    \Procedure{Get\_Sequence}{A,u,S,strokes\_cnt}\newline
    \Comment{%
      A is an adjacency list for N vertices. S is a list of
      strokes. strokes\_cnt keeps track of total number of
      strokes. \textbf{pop\_edge(u,w)} removes the edge (u,w) from the
      graph. A[u]\textbf{.get\_vertex()} returns a vertex from
      adjacency list of u.%
    }\\%
    \State If $A[u]$ $==$ $\mathbf{Null}$
    \State $\hspace{0.4cm}\text{\textbf{return}}$    
    \State $w \leftarrow$ $A[u]\textbf{.get\_vertex()}$
    \State $\textbf{pop\_edge(u,w)}$
    \State S[strokes\_cnt].\textbf{append(w)}
    \State $\Call{Get\_Sequence}{A,w,S,\mathit{strokes\_cnt}}$
    
    \EndProcedure 
  \end{algorithmic}
\end{algorithm}

\begin{algorithm}
  \caption{\texttt{Strokes\_Gen} Procedure for generating strokes from
    undirected graph}
  \label{algo:strokes_gen}
  \begin{algorithmic}[1]
  
    \Procedure{Strokes\_Gen}{A} \newline
    \Comment{Strokes\_Gen procedure called with adjacency list A}\\
    \State $\mathrm{strokes\_cnt}=0$
    \While {($A[v] \hspace{0.1cm}! \hspace{0.1cm}\mathbf{Null} \hspace{0.2cm}\forall v$)}
      \State S[strokes\_cnt].\textbf{append(v)}
      \State $u \leftarrow$ $A[v]\textbf{.get\_vertex()}$
      \State S[strokes\_cnt].\textbf{append(u)}
      \State $\textbf{pop\_edge(v,u)}$
      \State $\Call{Get\_Sequence}{A,u,S,\mathit{strokes\_cnt}}$
      \State $\text{strokes\_cnt}\gets{}\mathrm{strokes\_cnt} + 1$
    \EndWhile
      
    \EndProcedure 
  \end{algorithmic}
\end{algorithm}

The strokes are finally translated to
\textsc{gcode}\footnote{\textsc{gcode} is a standard numerical control
  programming language generally used as instructions to a
  \textsc{cnc}
  machine. Refer \url{https://en.wikipedia.org/wiki/G-code} for
  more information.} using a
simple translation method. Issue for each stroke in the set, the
following commands sequentially,
\begin{itemize}
\item push first vertex, (\texttt{Xff.ff Yff.ff}), 
\item engage (\texttt{G01 Z0}),
\item push rest of the vertices (\texttt{Xff.ff Yff.ff ...}), and
\item disengage (\texttt{G00 Z-5})
\end{itemize}

There are two caveats to remember here. Firstly, to disengage
(\texttt{G00 Z-5}) at the start. And secondly to rescale the
coordinates as per machine, for example, \texttt{Z-5} pertains to 5~mm away
from work, which in case of cm units or inch units, should be scaled by
an appropriate factor.

A very similar method is used to render to an SVG
Path~\cite{world_wide_web_consortium_w3c_paths_2018}, where each
stroke is a series of 2D points, expressed in XML as, \texttt{<path
  d="M x0 y0 L x1 y1 [L x y ...]" />} This is input to the robotic arm
to draw on paper.

\section{EXPERIMENTATION}
\label{sec:experimentation}
\subsection{Dataset}
\label{sec:dataset}

\begin{figure}[htbp]
  \centering
  \includegraphics[width=1.0\columnwidth]{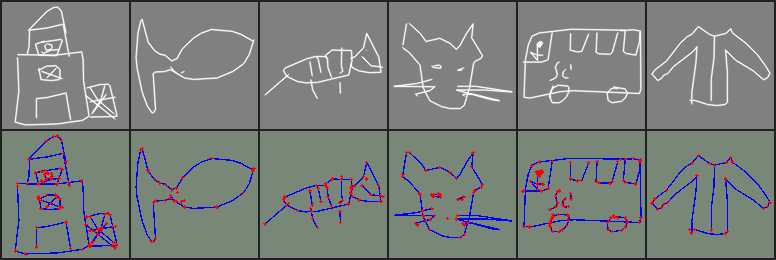}
  \caption{\textit{Top}. Samples from the \qdr{} dataset;
    \textit{Bottom}. Rendered true labels.}
  \label{fig:dataset}
\end{figure}

We use the \qdr{} dataset~\cite{ha_neural_2017} for the basic line
art. The sketches are available to us in the form of a list of
strokes, and each stroke is a sequence of points. The data is
normalised, centred to canvas, and re-scaled to accommodate for stroke
width, before being rasterised, using drawing functions, in order to
synthesise the input image, $X; X~\in~\mathbb{R}^{s\times{}s}$.

For the purpose of segmentation, an active pixel is either identified
as a corner (node in the graph), a connecting line, or the background;
i.e. $K=3$ classes. We compute the intersections using the vector data
from \qdr, and append them to the accumulated set of
corners, so that they are rasterised as $Y_{\mathrm{corners}}$. The
difference operation, with respect to the input, gives us the lines
channel, $Y_{\mathrm{lines}}=X - Y_{\mathrm{corners}}$. Everything
else is the background,
$Y_{\mathrm{bg}} = 1 - (Y_{\mathrm{corners}} +
Y_{\mathrm{lines}})$. Thus, we obtain the true labels,
$Y~\in~\mathbb{R}^{K\times{}s\times{}s}$. A glimpse of the dataset can
be seen in Fig.~\ref{fig:dataset}.

We created a set of $\sim{}65k$ samples for the training, and
$\sim{}10k$ samples for testing; each sample being a pair of images,
namely the input image and the true labels. We will make it available
to public domain after acceptance of this paper.

\subsection{Segmentation}
\label{sec:seg-experiment}

\begin{figure*}[htbp]
  \centering
  \includegraphics[width=\linewidth]{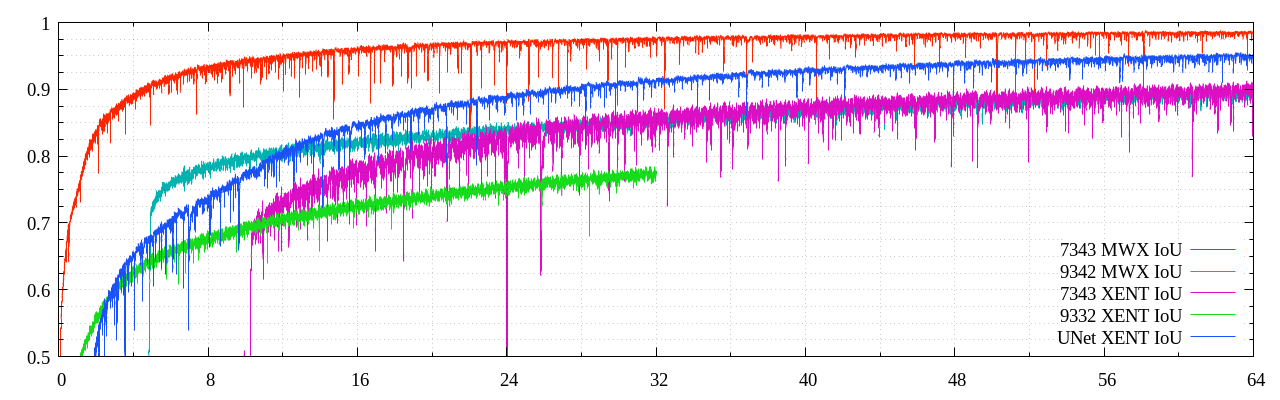}
  \caption{Validation Accuracy (\textsc{iou}'s) using models check-pointed
    during the training. (x-axis:\textit{Training epochs},
    y-axis:\textit{\textsc{iou}})}
  \label{fig:train-test}
\end{figure*}
\begin{figure*}[htbp]
  \centering
  \includegraphics[width=\linewidth]{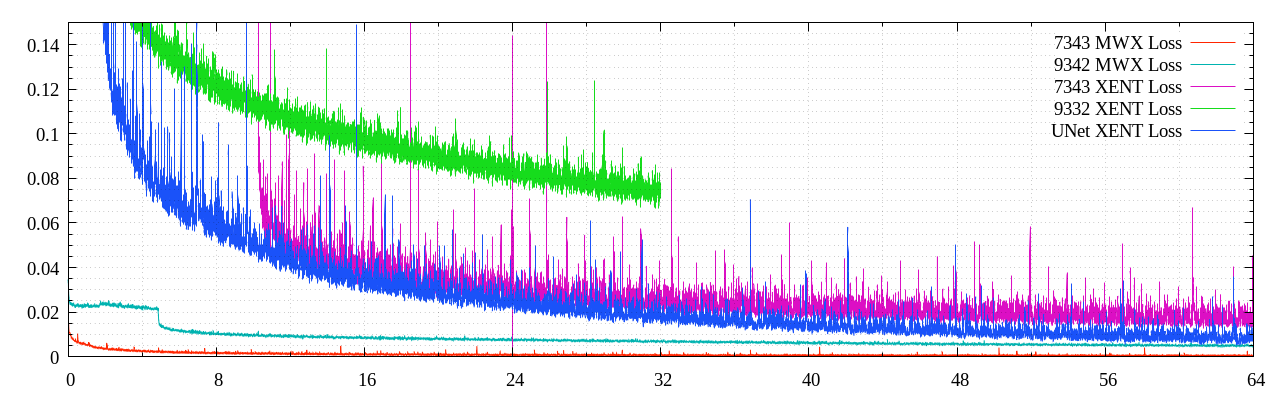}
  \caption{Validation Loss using models check-pointed during the
    training. (x-axis:\textit{Training epochs},
    y-axis:\textit{Loss})}
  \label{fig:train-loss}
\end{figure*}

We experimented with various combinations of network architectures,
and training objectives, to find a suitable segmentation network. The
most promising ones that have been presented here are: \textit{a)} Net
$(9\,3\,2\,2)$ with cross-entropy loss (\xent); \textit{b)} Net
$(9\,3\,3\,2)$ with \xent; \textit{c)} Net $(7\,3\,4\,3)$ with \xent;
\textit{d)} Net $(7\,3\,4\,3)$ with max-weighted cross-entropy loss
(\mwx). (For network nomenclature, refer \S~\ref{sec:inference} and
Fig.~\ref{network}.) Additionally, we also trained a variant of
vanilla U-Net with \xent, where we replaced `valid' convolutions with
padded-convolutions to obtain consistent image sizes on our
dataset. As per our nomenclature, the vanilla U-Net corresponds to
$(3\,3\,4\,2)$ architecture. The optimiser used was
RMSProp~\cite{Hinton_rmsprop}, with $\beta = 0.9$.

The validation accuracy and losses during the training are reported in
the Fig.~\ref{fig:train-test},\ref{fig:train-loss}. The x-axis
represents number of epochs of training in both the figures. The
y-axis represents the average \textsc{iou} (``intersection over
union'') in Fig.~\ref{fig:train-test}, and in
Fig.~\ref{fig:train-loss}, it represents the average loss. The model
was evaluated using validation set. We can see that net $(7\,3\,4\,3)$
with \mwx{} stabilises fastest and promises the most accurate
results. A quantitative summary of our evaluation over the test set,
is shown in the Table~\ref{tab:accuracy}. Here also we see a similar
trend, as expected, \ie $(7\,3\,4\,3)$ with \mwx{} is a winner by
far. Qualitatively, the implication of a better segmentation is most
prominently visible in fairly complicated cases, \eg
Fig.~\ref{fig:xent-effect}. This improvement in the results is
attributed to the design of max-weighted cross entropy loss (refer
\S~\ref{sec:loss_function}).

\begin{table}[htbp]
  
  \begin{center}
  \caption{Accuracy measure of segmentation networks}
  \label{tab:accuracy}
  \begin{tabular}{ccr}
    \hline
    Architecture & Loss Function & \textsc{iou} $(\%)$\\
    \hline
    Net $(7\,3\,4\,3)$ & \textsc{mwx} & \textbf{98.02\%}\\
    Net $(9\,3\,4\,2)$ & \textsc{mwx} & 91.24\%\\
    Net $(7\,3\,4\,3)$ & \textsc{xent} & 92.57\%\\
    Net $(9\,3\,3\,2)$ & \textsc{xent} & 75.09\%\\
    Vanilla U-Net  & \textsc{xent} & 94.23\%\\
    \hline
  \end{tabular}
  \end{center}
\end{table}

\subsection{Graph}
\label{sec:graph-experimentation}

\begin{figure}[hbtp]
  \centering
  \includegraphics[width=\columnwidth]{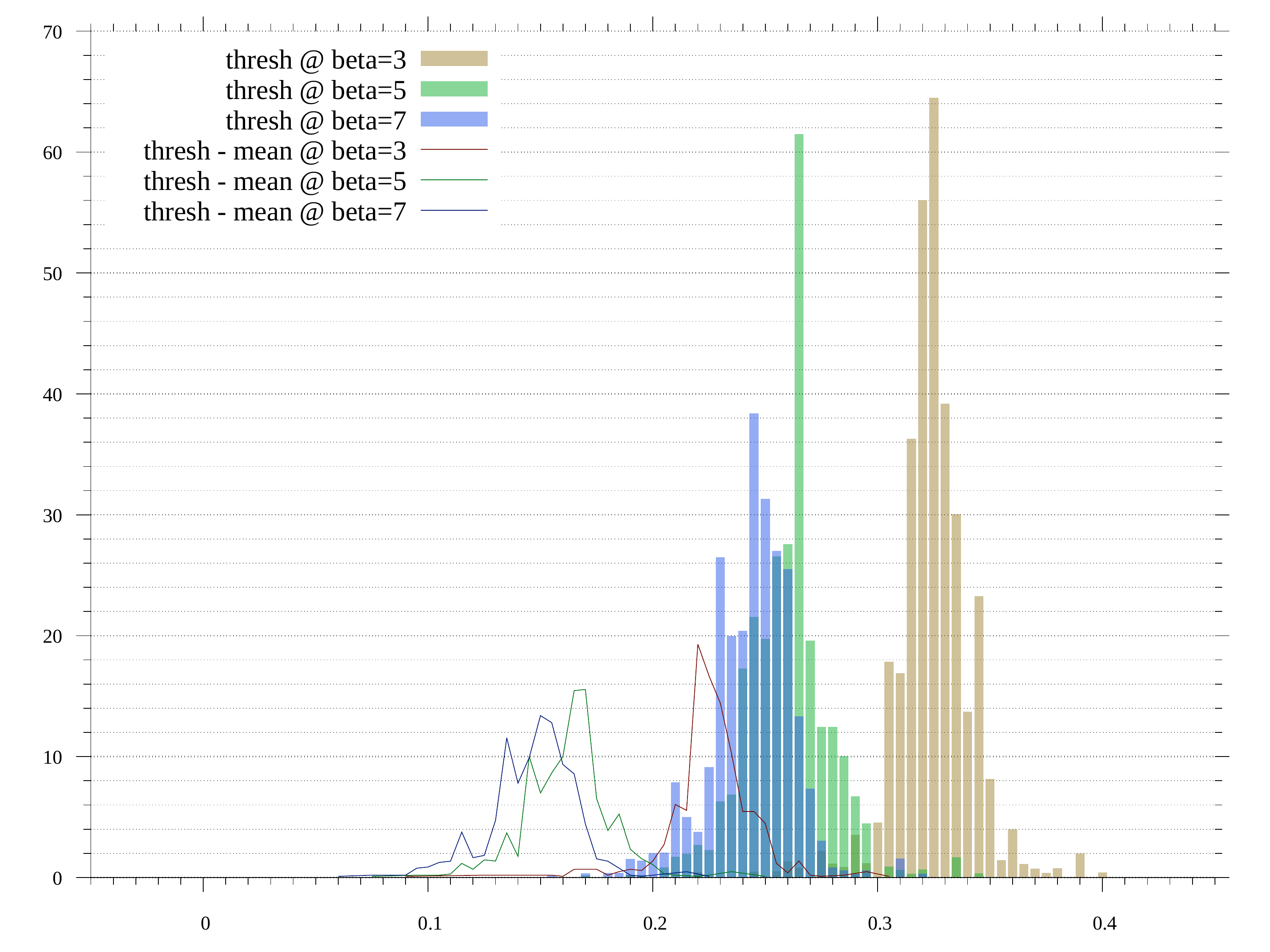}
  \caption{Histogram of approximate plausibility threshold
    $\widehat{\tau}$ and its distance from the average value of
    corresponding plausibility score $\overline{\eta}$ for different
    values of mask widths $\beta$. (x-axis:\textit{Threshold values})}
  \label{fig:pot-thresh}
\end{figure}

\begin{figure}[htbp]
  \centering
  \includegraphics[width=\columnwidth]{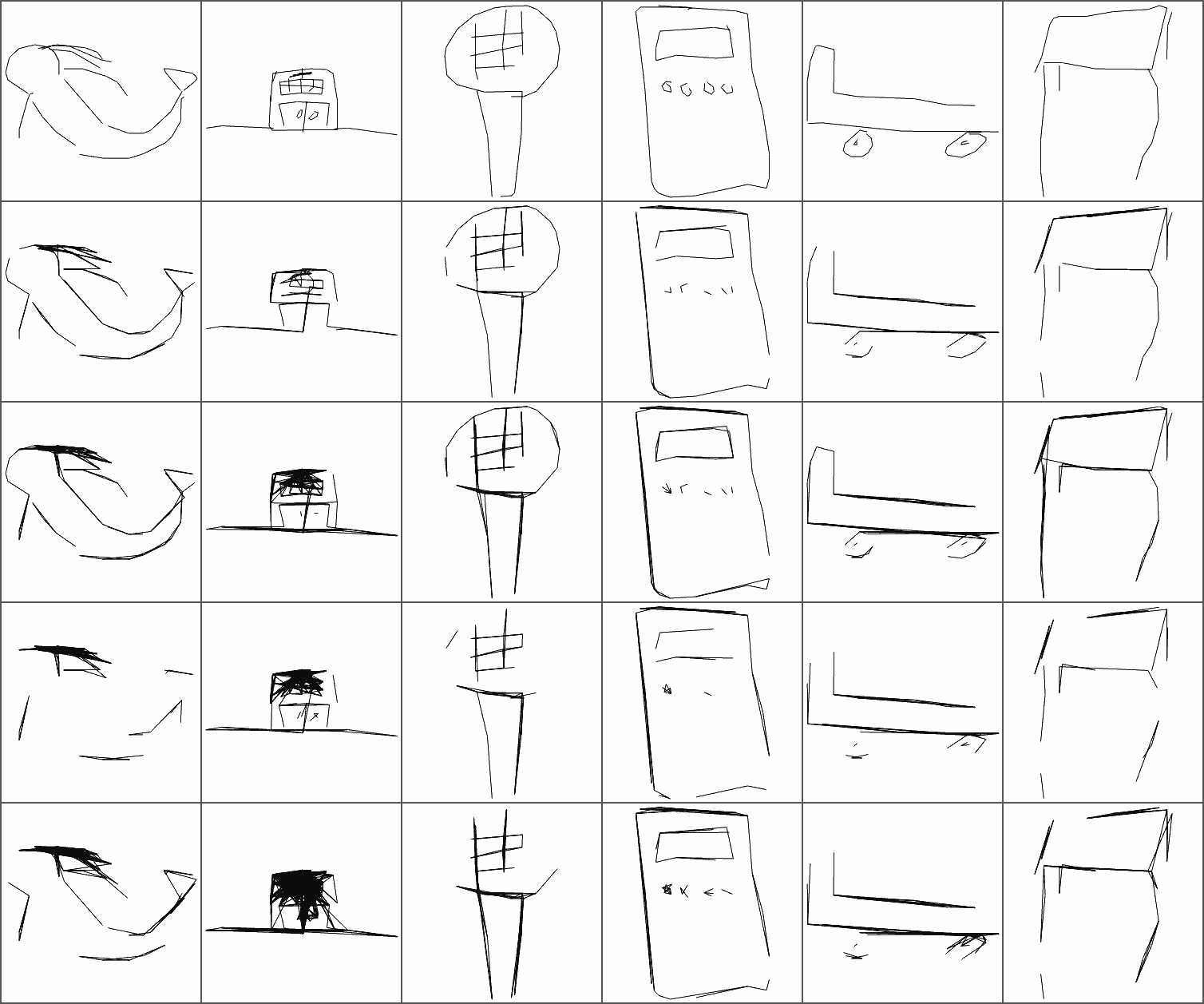}
  \caption{The top row shows the input. The subsequent rows show
    reconstruction using a graph interpreted from segmentation mask,
    with values of mask width and threshold $(\beta, \tau)$ set as (2,
    0.3), (3, 0.22), (5, 0.2), (7, 0.15) respectively downwards.}
  \label{fig:graph-interp-fixed-tau}
\end{figure}

\begin{figure}[htbp]
  \centering
  \includegraphics[width=\columnwidth]{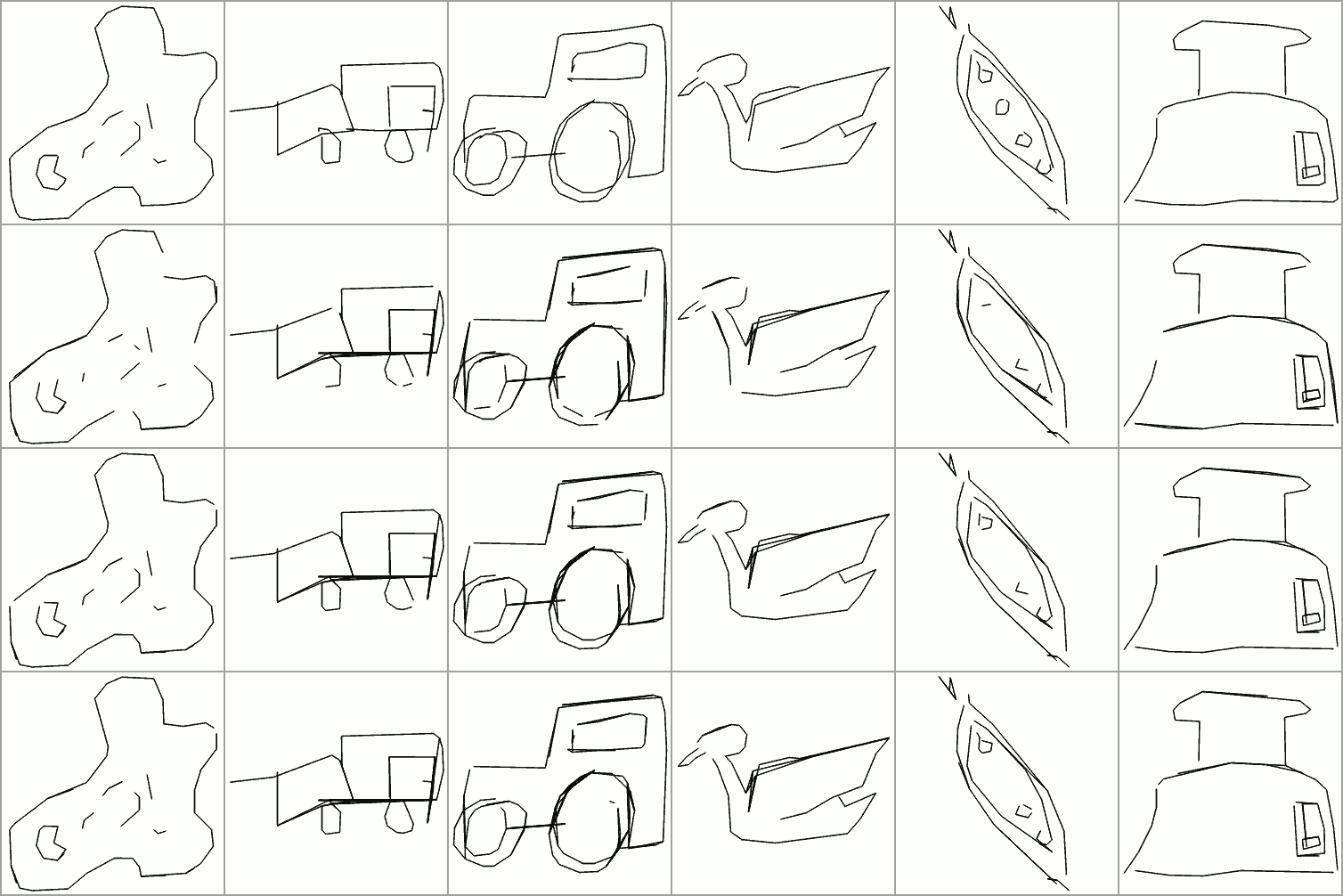}
  \caption{The results of feedback loop in graph
    interpretation. \textit{Rows top to bottom.} the input image;
    na\"{i}ve graph interpretation; interpretation after $N=5$
    updates; and after $N=10$ updates.}
  \label{fig:graph-fdbk}
\end{figure}

For graph interpretation, we analysed the distribution of plausibility
score over the test dataset, as defined in Eq.~\ref{eq:criterion},
with a view, that there should be a clear separation between the pairs
of centroids containing edges and the pairs without edges. To
investigate this, we experimented with three values of mask width
parameter $\beta \in \{3,\, 5,\, 7\}$. For a given image in the
dataset, we already know the number of lines, which we leverage to
find the approximate threshold $\widehat{\tau}$, by sorting the
plausibility scores $\eta_{\mathbf{p},\,\mathbf{q}}$ in reverse and
truncating. We smoothen out the value of $\widehat{\tau}$ by averaging
over a neighbourhood of size $k=3$.  Further, we collect the values of
$\widehat{\tau}$ and average plausibility score $\overline{\eta}$ for
selected values of $\beta$, across the images in test set. The
histograms for $\widehat{\tau}$ in Fig~\ref{fig:pot-thresh} show a
single prominent peak and light tail; and the histograms of
$\widehat{\tau} - \overline{\eta}$ show that there is a comfortable
margin between $\widehat{\tau}$ and $\tau$, which implies that, for
our distribution from \qdr{} dataset, there indeed \textit{is a clear
  separation between the pairs of centroids containing edges and the
  pairs without edges.}

Further we qualitatively tested the method with values of mask-width
$\beta$ and a fixed plausibility threshold of $\tau$ set to
$(2,\, 0.3)$, $(3,\, 0.22)$, $(5,\, 0.2)$ and $(7,\, 0.15)$
respectively. The results are shown in
Fig.~\ref{fig:graph-interp-fixed-tau}. We see that a fixed value of
threshold is not able to capture the edges based on the plausibility
score.

Thereafter, we tested the adaptive update method, designed in
Eq.~\ref{eq:update}, with heuristically set parameters: mask width
$\beta=1.8$, initial plausibility threshold $\tau^{(0)}=0.35$, update
rate $\lambda=0.05$. Typically, we found satisfactory results with
$N=10$ updates. In Fig.~\ref{fig:graph-fdbk}, the second row is a
rendering of graph interpreted with $\tau^{(0)}$, and has missed a few
strokes. In almost all the cases after $N=10$ updates (bottom row) the
missing strokes have been fully recovered!

\subsection{Application}
\label{sec:application}

We tested the results for application with two robots. First is a
traditional \textsc{cnc} milling machine ``\textsc{emco} Concept Mill
250,'' where we used pen as a tool to plot over a plane surface. This
machine accepts \textsc{gcode} as input. Second is a modern robotic
arm ``Dobot Magician Robotic Arm,'' that holds a pen to draw on
paper. This machine accepts line drawings in \textsc{svg} format, such
that each stroke is defined as a \textit{path} element. The method of
post processing the list of strokes to either create \textsc{gcode}
for \textsc{cnc} plotter or create \textsc{svg} for a robotic arm is
detailed out in \S~\ref{sec:graph-partition}. For practical purposes,
we fixed the drawings in either case to be bounded within a box of
size $64 \mathrm{mm} \times 64 \mathrm{mm}$ and set the origin at
$(25 \mathrm{mm},\, 25 \mathrm{mm})$ Fig.~\ref{fig:cncvsrobo} shows a
still photograph of the robots in action. Reader is encouraged to
watch the supplementary video for a better demonstration.

\subsection{Generalisation}
\label{sec:generalization}

\begin{figure}[htbp]
  \centering
  \includegraphics[width=\columnwidth]{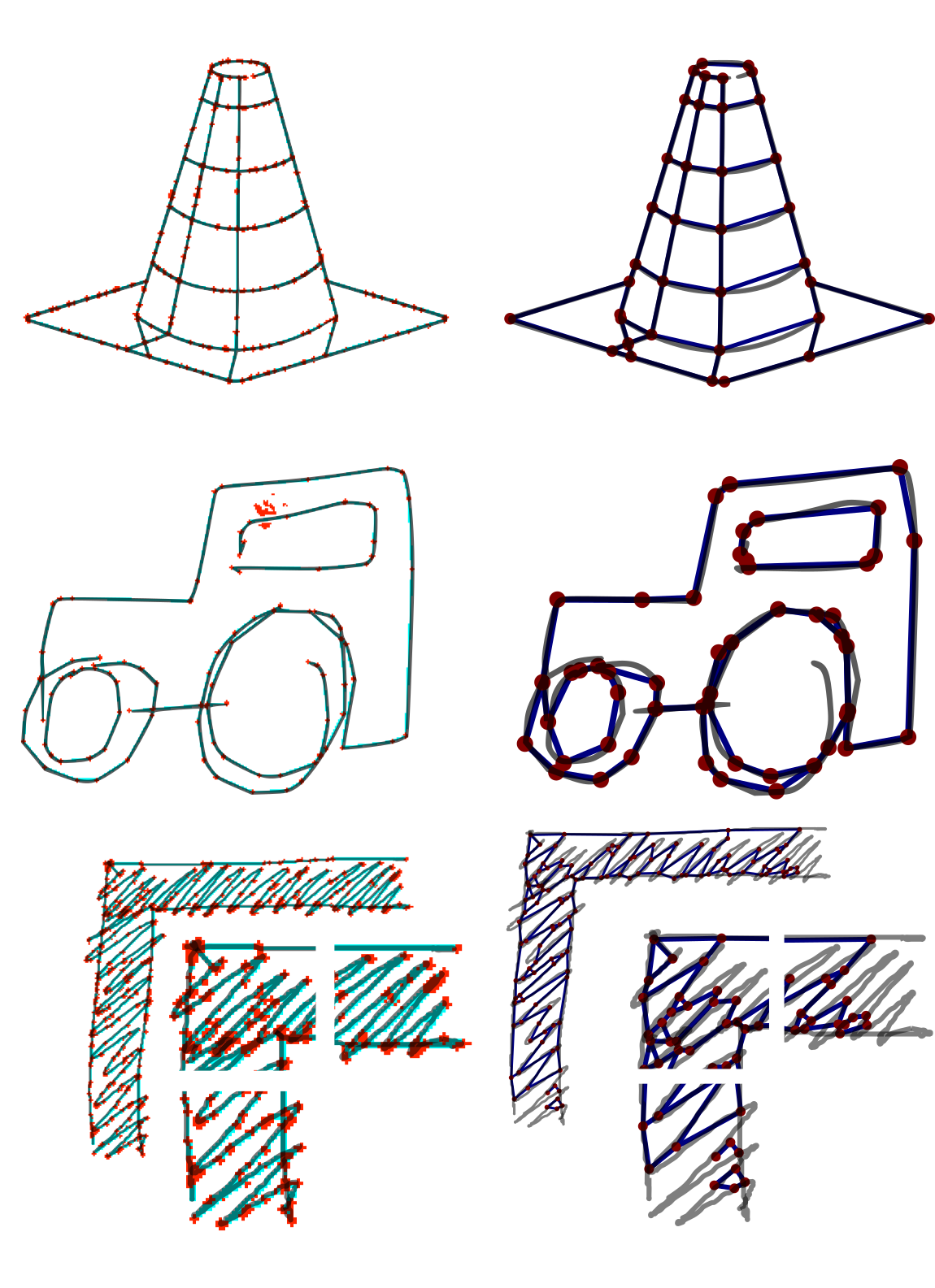}
  \caption{Comparison of our segmentation (left column) against
    Favreau's vectorization (right column). \textit{Top to Bottom.} A
    simple line sketch from Favreau's examples; A simple line sketch
    from \textit{Quick-draw} dataset; A complicated hatch pattern with
    zoomed in details.}
  \label{fig:favreau-comparison}
\end{figure}

A deep learning based method is generally preferred for it generalises
well. Here we qualitatively inspect our model against Favreau \etal's
vectorization method\cite{favreau2016fidelity} to line drawing
vectorization, using three different samples, namely a simple sketch
from their samples; a simple sketch from \qdr; and a complicated
sketch, representing a hatch pattern, commonly used to represent walls
in architectural graphics. In Fig.~\ref{fig:favreau-comparison}, we
see that our method produces satisfactory segmentation on both the
simple cases, and also gracefully degrades its performance in the
complicated case; whereas, the latter method, compromises on details,
\eg few open strokes near the wheels of the automobile, and thus fails
miserably on the complicated task of vectorizing a hatch pattern,
missing out several details.

\section{CONCLUSION}

We have portrayed an alternate method for a drawing robot in the
context of line drawings, which maps image domain to a graph structure
that represents the original image, and retrieve a set of actionable
sequences.

There is definitely, a scope of improvement here. We see the following
points to be most significant,
\begin{itemize}
\item The segmentation model is highly sensitive to noise, and similar
  phenomenon has been observed in the context of deep learning by
  Goodfellow et.al~\cite{goodfellow2014explaining}. This problem is
  subject of contemporary research.
\item Graph interpretation method is sensitive to initial choice of
  threshold parameter $\tau$ and the update hyperparameter $\lambda$,
  refer eq.~\ref{eq:update}.
\end{itemize}

\begin{figure}[hbtp]
  \centering
  \includegraphics[width=\columnwidth]{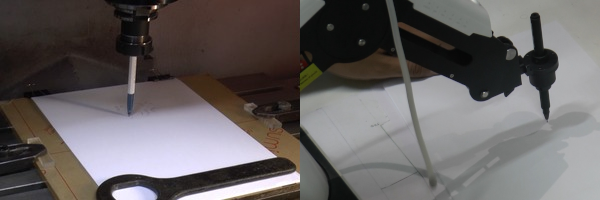}
  \caption{Demonstration of the applicability of our output using two
    different machines, namely, the \textsc{cnc} plotter in the left,
    and the robotic arm in the right.}
  \label{fig:cncvsrobo}
\end{figure}

The pipeline proposed by us is a multi utility technique, the use of
which we have portrayed through a CNC machine as well as through a
robotic arm as shown in Fig.~\ref{fig:cncvsrobo}

{\small
\bibliographystyle{ieee}
\bibliography{main}
}

\end{document}